\documentclass{article}

\usepackage[preprint]{neurips_2026}

\usepackage[utf8]{inputenc} 
\usepackage[T1]{fontenc}    
\usepackage{microtype}
\usepackage{graphicx} 
\usepackage{subfig}
\usepackage{booktabs} 
\usepackage{hyperref}       
\usepackage{algorithm}

\usepackage{wrapfig}
\usepackage{bbding}

\usepackage[T1]{fontenc}    
\usepackage{url}            
\usepackage{amsfonts}       
\usepackage{nicefrac}       
\usepackage{xcolor}         

\usepackage{xspace}
\usepackage{textcomp}

\usepackage{stmaryrd}

\usepackage{amsmath}
\usepackage{amssymb}
\usepackage{mathtools}
\usepackage{amsthm}
\usepackage{dsfont}
\usepackage{diagbox}
\usepackage{adjustbox}
\usepackage{multirow}
\usepackage{makecell}
\usepackage{colortbl}
\usepackage{tikz}
\usepackage{bm}

\definecolor{cvprblue}{rgb}{0.21,0.49,0.74}
\definecolor{mygrey}{rgb}{ .863,  .863,  .863}
\definecolor{mypink}{rgb}{0.999,  0.414,   0.414}
\definecolor{mygreen}{rgb}{ .612,  .859,  .659}
\definecolor{myred}{rgb}{ .906,  .600,  .569}

\usepackage[capitalize,noabbrev]{cleveref}

\theoremstyle{plain}
\newtheorem{theorem}{Theorem}

\theoremstyle{definition}

\theoremstyle{remark}
\newtheorem*{remark}{Remark}

\usepackage[textsize=tiny]{todonotes}

\usepackage[T1]{fontenc}    
\usepackage{url}            
\usepackage{amsfonts}       
\usepackage{nicefrac}       
\usepackage{xcolor}         

\title{Delve into the Applicability of Advanced Optimizers for Multi-Task Learning}

\author{
Zhipeng Zhou$^{1}$, \quad Linxiao Cao$^{2}$, \quad Pengcheng Wu$^{1}$, \quad Peilin Zhao$^{3}$, \quad Chunyan Miao$^{1}$\\
$^{1}$Nanyang Technological University \\
$^{2}$Hong Kong University of Science and Technology (Guangzhou) \\
$^{3}$School of Artificial Intelligence, Shanghai Jiao Tong University\\
{\ttfamily\small zzpustcml@gmail.com, lcao950@connect.hkust-gz.edu.cn, peilinzhao@sjtu.edu.cn} \\
{\ttfamily\small \{pengchengwu, ascymiao\}@ntu.edu.sg}
}

\begin{document}

\maketitle

\begin{abstract}
Multi-Task Learning (MTL) is a foundational machine learning problem that has seen extensive development over the past decade. Recently, various optimization-based MTL approaches have been proposed to learn multiple tasks simultaneously by altering the optimization trajectory. Although these methods strive to de-conflict and re-balance tasks, we empirically identify that their effectiveness is often undermined by an overlooked factor when employing advanced optimizers: the instant-derived gradients play only a marginal role in the actual parameter updates. This discrepancy prevents MTL frameworks from fully releasing its power on learning dynamics. Furthermore, we observe that Muon—a recently emerged advanced optimizer—inherently functions as a multi-task learner, which underscores the critical importance of the gradients used for its orthogonalization. To address these issues, we propose \texttt{APT} (Applicability of advanced oPTimizers), a framework featuring a simple adaptive momentum mechanism designed to balance the strengths between advanced optimizers and MTL. Additionally, we introduce a light direction preservation method to facilitate Muon’s orthogonalization. Extensive experiments across four mainstream MTL datasets demonstrate that \texttt{APT} consistently augments existing MTL approaches, yielding substantial performance improvements.
\end{abstract}

\section{Introduction}
Traditional machine learning typically relies on customized models tailored with task-specific knowledge. However, in modern real-world scenarios, the requirement to handle multiple tasks simultaneously has made the development of Multi-Task Learning (MTL) models increasingly essential.

MTL is a mature field with diverse methodologies, including architecture-based approaches~\cite{zhao2024layer,heuer2021multitask}, optimization-based methods~\cite{liu2021conflict,liu2023famo}, and task relationship modeling~\cite{song2022efficient,fifty2021efficiently}. This paper focuses exclusively on optimization-based MTL, which assumes a fixed architecture comprising a shared backbone and task-specific branches~\cite{liu2019end}. These methods aim to design effective re-weighting strategies for task gradients or losses.

\begin{wrapfigure}[19]{r}{0.5\textwidth}
    \vspace*{-\baselineskip}
    \centering
    \includegraphics[width=0.9\linewidth]{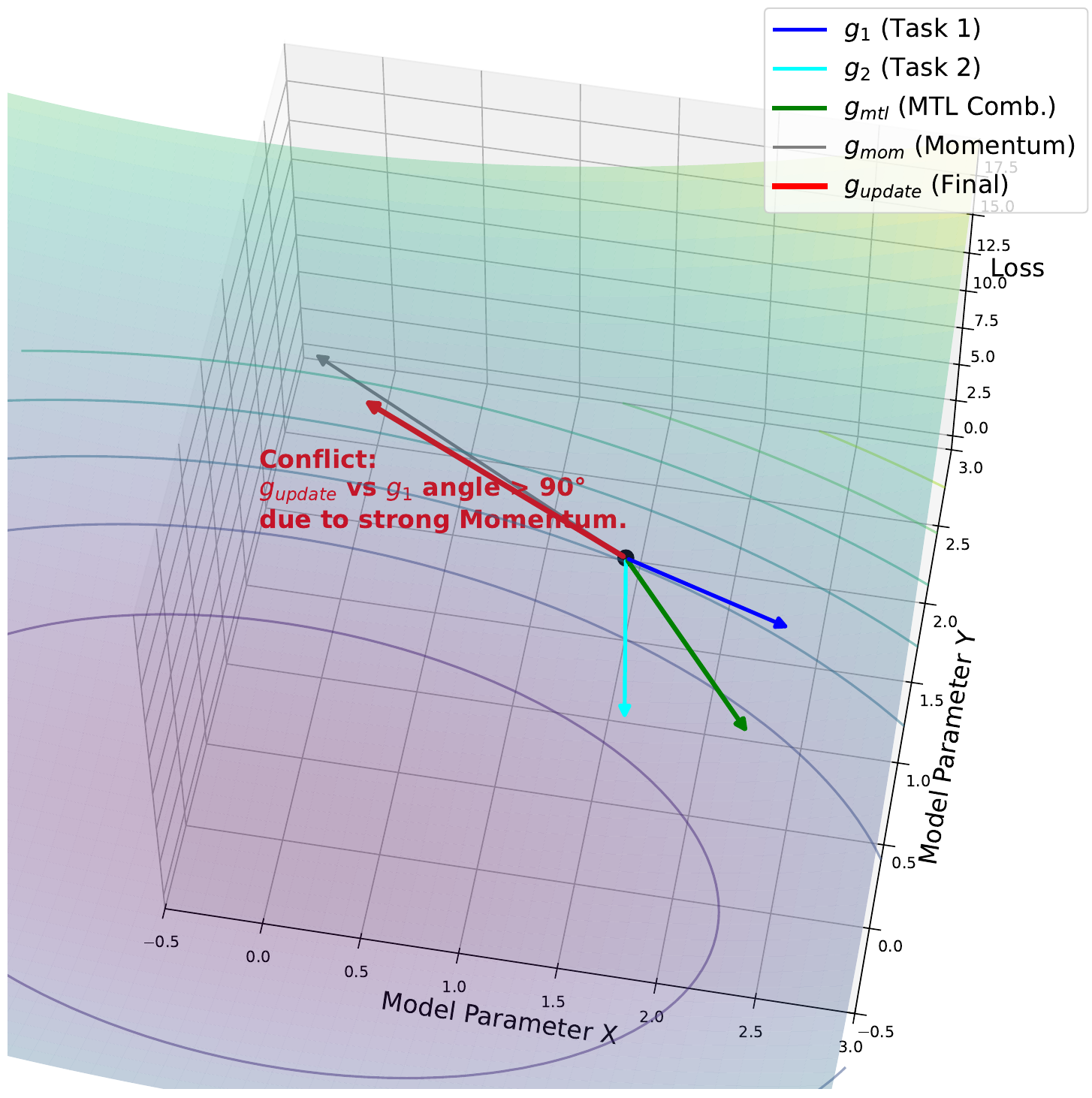}
    \caption{Illustration of the impact of EMA in advanced optimizer for MTL. Here we adopt $g_{update} = 0.9 * g_{mom} + 0.1 * g_{mtl}$ as the example. } 
    \label{fig:illu}
\end{wrapfigure}

Recent benchmarks~\cite{ban2024fair,zhou2025continual} demonstrate that advanced optimization-based MTL achieves competitive performance, sometimes even surpassing Single-Task Learning (STL) upper bounds. Nevertheless, we contend that the effectiveness of these approaches is fundamentally undermined by the use of Exponential Moving Average (EMA) in modern optimizers, e.g., Adam series~\cite{kingma2014adam,loshchilov2017decoupled}, and Muon~\cite{jordan2024muon}, etc. As illustrated in Figure~\ref{fig:illu}, even when a de-conflicted direction $g_{mtl}$ is derived, the EMA mechanism causes the momentum direction $g_{mom}$ to dominate, resulting in a final update $g_{update}$ that remains conflicted. This suggests that current MTL mechanisms do not achieve the \textbf{instant de-conflicting} intended; instead, they \textbf{amortize} MTL effects over time. This claim is further supported by our empirical observations in Section~\ref{sec:mtl_not_work}. While EMA is beneficial for noise reduction and stability, its use creates a critical trade-off between instant and amortized de-conflicting (\textbf{\textit{Challenge 1}}). In this work, we focus on Adam, AdamW, and Muon as representative advanced optimizers for our study.

Furthermore, Muon—a recently proposed advanced optimizer—requires additional consideration. We show that Muon can be interpreted as an implicit MTL learner, as its orthogonalization operation tends to amplify components shared among task gradients. This property places stricter requirements on the quality of the aggregated input gradient, giving rise to a second challenge (\textbf{\textit{Challenge 2}}).


To address these challenges, we propose a simple adaptive momentum strategy that dynamically balances instant and amortized de-conflicting based on local curvature. To resolve \textit{Challenge 2}, we additionally introduce a light direction preservation approach designed to facilitate Muon’s orthogonalization process. In a nutshell, our contributions can be summarized into three-fold:
\begin{itemize}
    \item We reveal that mainstream optimization-based MTL approaches primarily achieve amortized rather than instant de-conflicting, which may deviate from common expectations and partially explains their limitations in practice.
    \item To jointly benefit from the denoising and smoothing effects of EMA and the de-conflicting capability of MTL, we propose an adaptive momentum strategy guided by curvature information. In addition, we introduce a lightweight direction preservation (LDP) approach to derive a more favorable input gradient for the Muon optimizer.
    \item Extensive experiments across multiple mainstream MTL benchmarks have demonstrated that our method successfully augments mainstream MTL approaches, achieving a new competitive state-of-the-art (SOTA) performance.  
\end{itemize}
\section{Related Work}

\subsection{Optimization-based Multi-Task Learning}
In optimization-based MTL, a wide range of methods have been proposed~\cite{zhouexploring,zhou2025injecting}, which can be broadly categorized into gradient-based and loss-based approaches. Gradient-based methods operate directly on task gradients with respect to shared parameters and aim to resolve conflicts through gradient manipulation. For example, MGDA~\cite{sener2018multi} seeks a convex combination of task gradients with minimum norm. PCGrad~\cite{yu2020gradient} addresses gradient conflicts by projecting conflicting gradients onto orthogonal subspaces. CAGrad~\cite{liu2021conflict} balances minimum-norm optimization with convergence guarantees. Align-MTL~\cite{senushkin2023independent} performs singular value decomposition (SVD) on task gradients and re-weights gradient components in the resulting subspace. Nash-MTL~\cite{navon2022multi} and FairGrad~\cite{ban2024fair} both formulate MTL as a fair allocation problem across tasks, with Nash-MTL being a special case under certain settings. COST~\cite{zhou2025continual} tackles gradient conflicts by seeking alternative update directions that lie on the same loss level set.

In contrast, loss-based methods adjust task weights based solely on task losses computed from model outputs, without explicitly using task gradients. As a result, they are typically more lightweight but often underperform gradient-based approaches due to the lack of gradient-level information. Early loss-based methods include RLW~\cite{lin2021reasonable}, which samples task weights from a random distribution, and Uncertainty Weighting~\cite{kendall2018multi}, which derives weights based on task-specific predictive uncertainty. More recent loss-based approaches are commonly formulated within a bi-level optimization framework, differing mainly in their lower-level task-weight update strategies. For instance, FAMO~\cite{liu2023famo} amortizes the MGDA objective over time using a first-order approximation. Go4Align~\cite{shen2024go4align} introduces dynamic task grouping in the lower-level optimization. LDC-MTL~\cite{xiao2025ldc} employs a routing network to regulate loss discrepancies during training.

\begin{table*}[ht]
\setlength\tabcolsep{4pt}
\centering
\caption{Analysis on $\beta_1$ across various MTL approaches and advanced optimizers on \textit{CityScapes}. AdamW is set with weight decay as 1.e-4 to distinguish with Adam. All results are average on 3 random seeds. The best scores are provided in \colorbox{gray!20}{gray}.}
\label{tab:pre_results}
\begin{tabular}{l|c c c | c c c }
\hline \toprule 
\multirow{2}{*}{$\beta_1$} & \multicolumn{3}{c|}{MGDA} & \multicolumn{3}{c}{CAGrad}\\
\cmidrule{2-7}
 & Adam & AdamW & Muon & Adam & AdamW & Muon \\
\midrule
default & 7.29 {\footnotesize $\pm$ 4.61} & 6.41 {\footnotesize $\pm$ 5.71}  & \cellcolor{mygrey}3.37 {\footnotesize $\pm$ 4.02} & 12.38 {\footnotesize $\pm$ 1.49} & 10.57 {\footnotesize $\pm$ 1.80} & \cellcolor{mygrey}-2.37 {\footnotesize $\pm$ 0.54}  \\
0.6     & 9.21 {\footnotesize $\pm$ 4.12} & 5.82 {\footnotesize $\pm$ 3.37}  & 4.14 {\footnotesize $\pm$ 1.10} & 9.78 {\footnotesize $\pm$ 1.14} & 10.88 {\footnotesize $\pm$ 1.54} & 1.95 {\footnotesize $\pm$ 0.90}  \\
0.2     & 9.25 {\footnotesize $\pm$ 3.75} & \cellcolor{mygrey}5.17 {\footnotesize $\pm$ 3.24}  & 19.39 {\footnotesize $\pm$ 6.64} & 10.78 {\footnotesize $\pm$ 3.03} & \cellcolor{mygrey}10.07 {\footnotesize $\pm$ 0.96} & 81.12 {\footnotesize $\pm$ 9.13} \\
0.0     & \cellcolor{mygrey}3.52 {\footnotesize $\pm$ 1.50} & 10.23 {\footnotesize $\pm$ 4.68}  & 33.35 {\footnotesize $\pm$ 6.61} & \cellcolor{mygrey}9.11 {\footnotesize $\pm$ 0.97}  & 10.08 {\footnotesize $\pm$ 2.58} & 120.78 {\footnotesize $\pm$ 36.70} \\
\midrule
\multirow{2}{*}{$\beta_1$} & \multicolumn{3}{c|}{Nash-MTL} & \multicolumn{3}{c}{FairGrad}\\
\cmidrule{2-7}
 & Adam & AdamW & Muon & Adam & AdamW & Muon \\
\midrule
default & 24.87 {\footnotesize $\pm$ 1.57} & 23.42 {\footnotesize $\pm$ 1.61}  & \cellcolor{mygrey}-4.22 {\footnotesize $\pm$ 0.66} & 4.23 {\footnotesize $\pm$ 3.20} & 4.04 {\footnotesize $\pm$ 0.66} & 6.04 {\footnotesize $\pm$ 1.16} \\
0.6     & 22.59 {\footnotesize $\pm$ 1.39} & 21.62 {\footnotesize $\pm$ 1.89}  & 3.15 {\footnotesize $\pm$ 2.97} & 4.81 {\footnotesize $\pm$ 2.77} & 3.53 {\footnotesize $\pm$ 4.19} & \cellcolor{mygrey}1.63 {\footnotesize $\pm$ 3.05}  \\
0.2     & \cellcolor{mygrey}20.45 {\footnotesize $\pm$ 1.68} & \cellcolor{mygrey}19.05 {\footnotesize $\pm$ 2.30}  & 77.04 {\footnotesize $\pm$ 27.68} & 2.86 {\footnotesize $\pm$ 1.13} & \cellcolor{mygrey}2.94 {\footnotesize $\pm$ 0.88} & 44.29 {\footnotesize $\pm$ 25.49}  \\
0.0     & 20.68 {\footnotesize $\pm$ 2.04} &  19.68 {\footnotesize $\pm$ 0.99} & 92.93 {\footnotesize $\pm$ 5.26} & \cellcolor{mygrey}2.38 {\footnotesize $\pm$ 0.90} & 3.26 {\footnotesize $\pm$ 1.22} & 55.87 {\footnotesize $\pm$ 28.48} \\
\hline \toprule 
\end{tabular}
\end{table*}
\subsection{Advanced Optimizers}
Advanced optimizers in deep learning are dominated by the Adam family, which builds on exponential moving averages of first- and second-order moments to provide per-parameter adaptive steps~\cite{kingma2014adam}. Subsequent variants target stability, generalization, and theoretical issues: AdaMax replaces the second-moment normalization with an $\ell_\infty$-based scheme; AMSGrad~\cite{reddi2019convergence} introduces a non-increasing second-moment accumulator to address non-convergence behaviors in Adam-type methods; AdamW decouples weight decay from the adaptive update, improving regularization and often empirical generalization~\cite{loshchilov2017decoupled}; RAdam~\cite{liu2019variance} rectifies the early-stage variance of the adaptive learning rate, reducing sensitivity to warmup heuristics; and AdaBound~\cite{luo2019adaptive} dynamically bounds learning rates to interpolate between adaptive methods and SGD. More recently, Muon~\cite{jordan2024muon} departs from element-wise adaptivity by applying matrix orthogonalization to parameter updates (notably for hidden-layer weight matrices), aiming to improve conditioning and achieve higher compute-efficiency in large-scale pretraining.

\noindent \textbf{Discussion with Counterparts.} To the best of our knowledge, UPGrad~\cite{quinton2024jacobian} is the most related work with this paper. In its implementation, it adopts both SGD and Adam optimziers to demonstrate the compatibility of the proposed method, but fails to provide more insights on this perspective. 

\section{Motivation}

\subsection{Preliminary}
Advanced optimizers typically share a common momentum mechanism; here, we use Adam as a representative example to illustrate this process. The momentum update is defined as:
\begin{align} \label{eqn:momentum}
    m_t = \beta_1 * m_{t-1} + (1 - \beta_1) * G_t
\end{align}
where $m_t$ represents the momentum (the first moment vector) at time step $t$. $\beta_1$ is a hyperparameter, typically set to $0.9$ or higher. $G_t$ is the gradient derived from the current step. In the context of MTL, $G_t$ is the aggregated gradient produced by a specific MTL algorithm (e.g., MGDA, CAGrad, or Nash-MTL, etc). In the following, we first reveal a fundamental mismatch between the momentum mechanism used in advanced optimizers and the objectives of MTL. We then further discuss the applicability and unique characteristics of the Muon optimizer in the MTL setting.
\begin{figure*}
    \centering
    \subfloat[Similarity over Ave. (LS).]{\includegraphics[width = 0.245\textwidth]{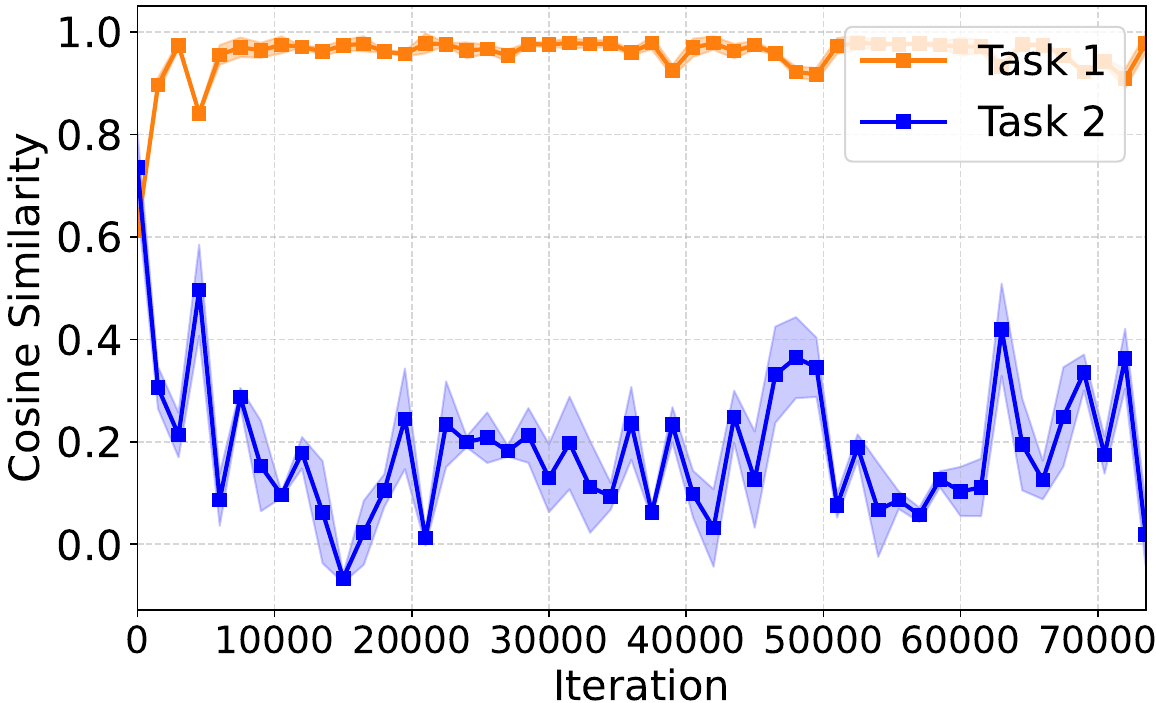}}
    \subfloat[Similarity over FairGrad.]{\includegraphics[width = 0.245\textwidth]{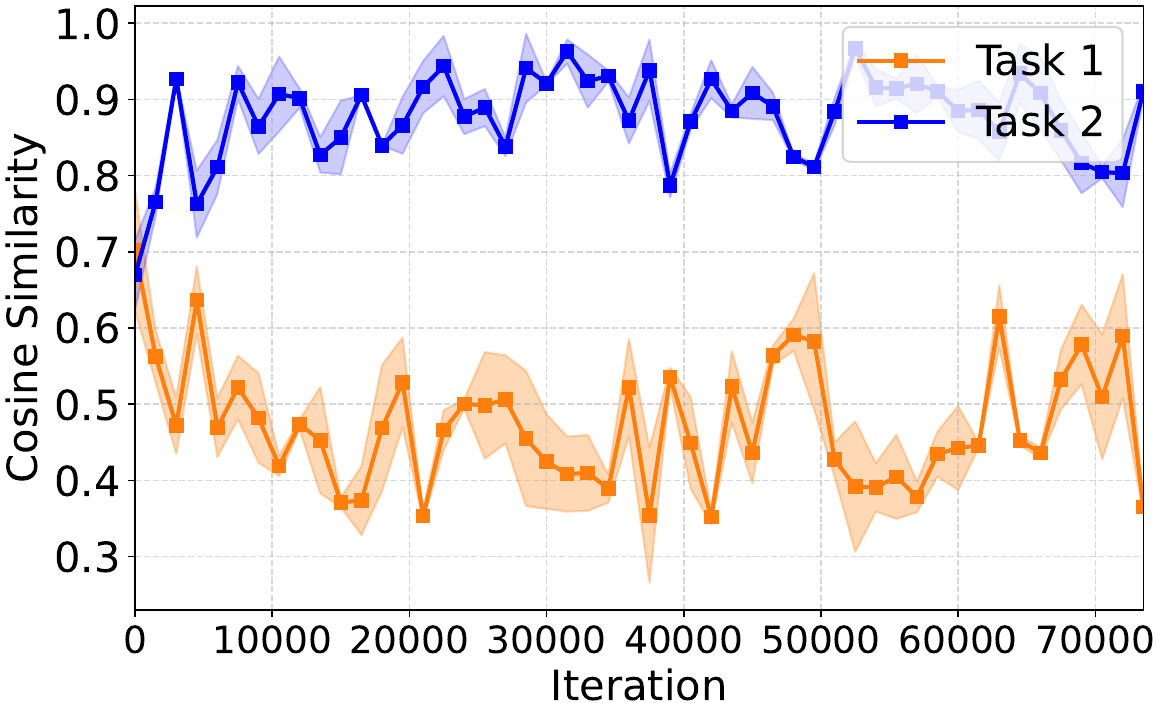}}
    \subfloat[Similarity over Momentum.]{\includegraphics[width = 0.245\textwidth]{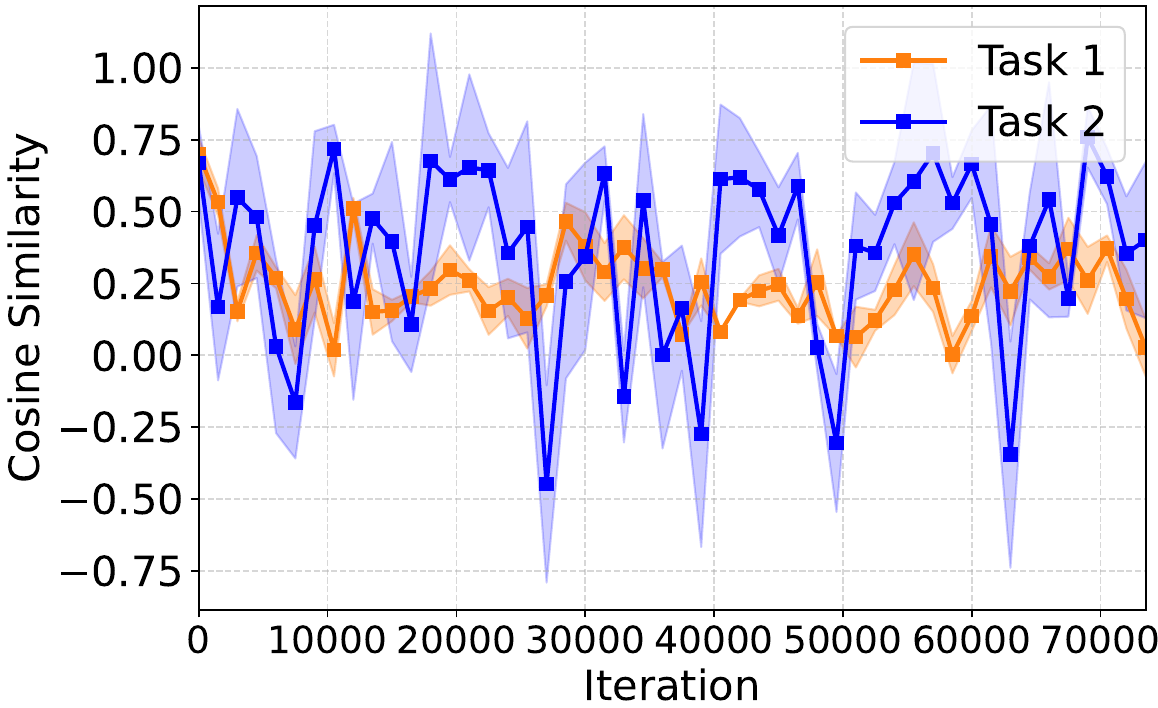}}
    \subfloat[Project Norm Comparison.]{\includegraphics[width = 0.245\textwidth]{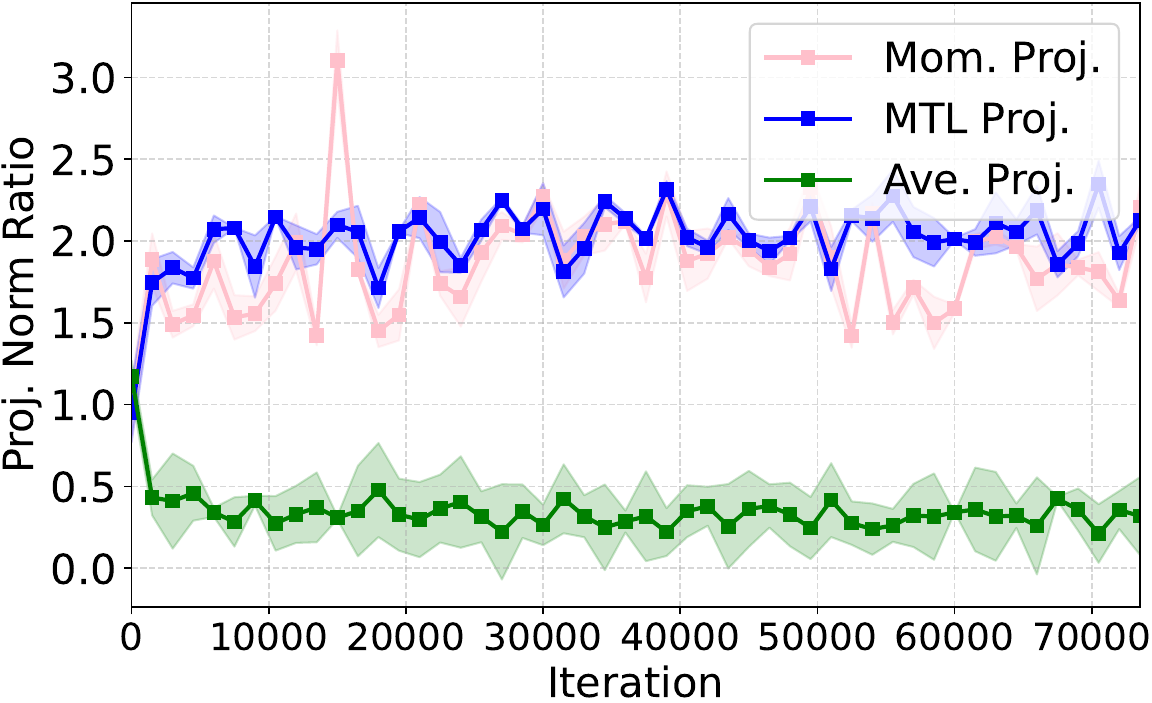}}    
    \caption{Similarity and project norm comparison.} 
    \label{fig:sim_proj}
\end{figure*}
\begin{figure*}
    \centering
    \subfloat[MGDA.]{\includegraphics[width = 0.245\textwidth]{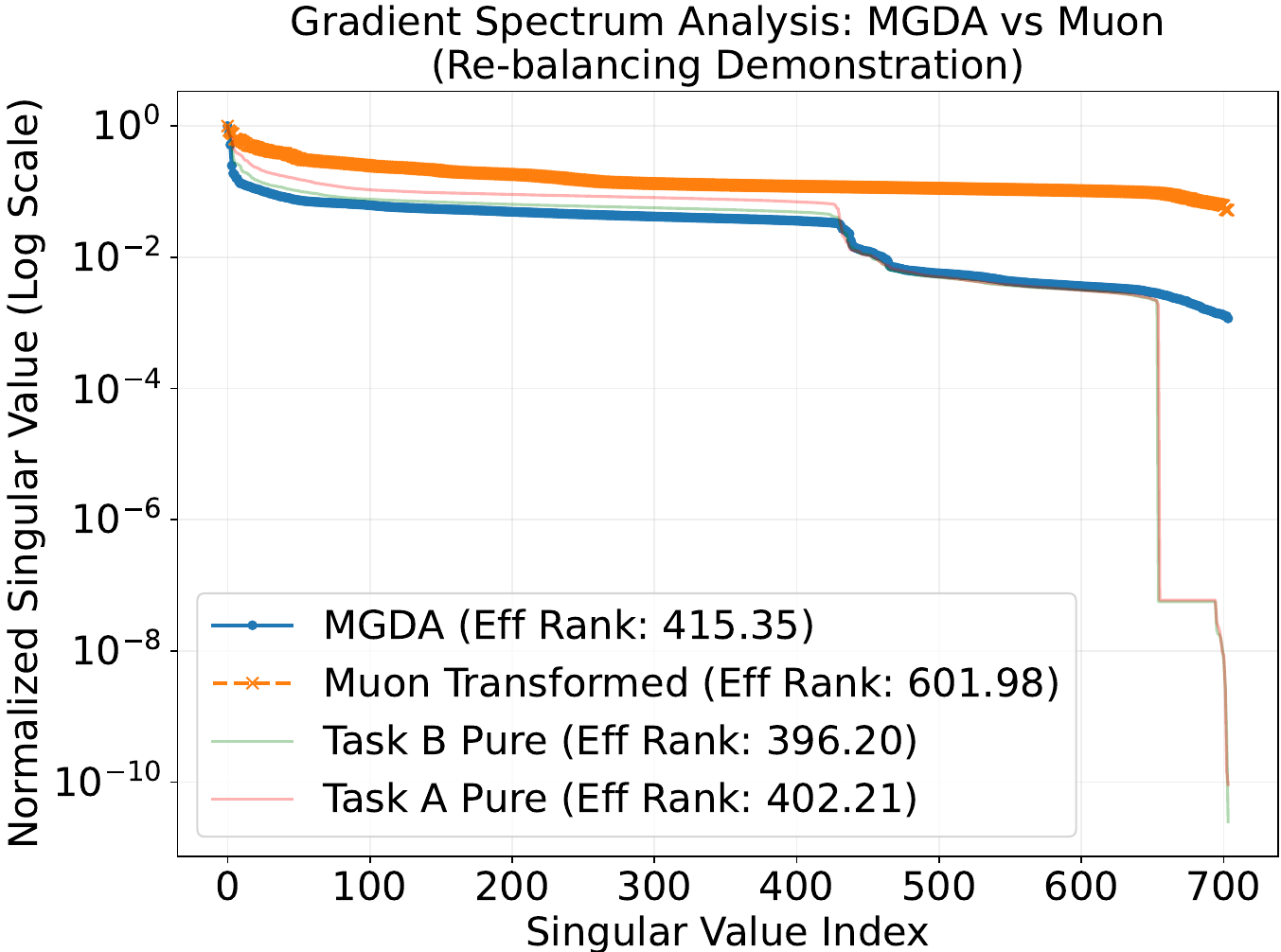}}
    \subfloat[CAGrad.]{\includegraphics[width = 0.245\textwidth]{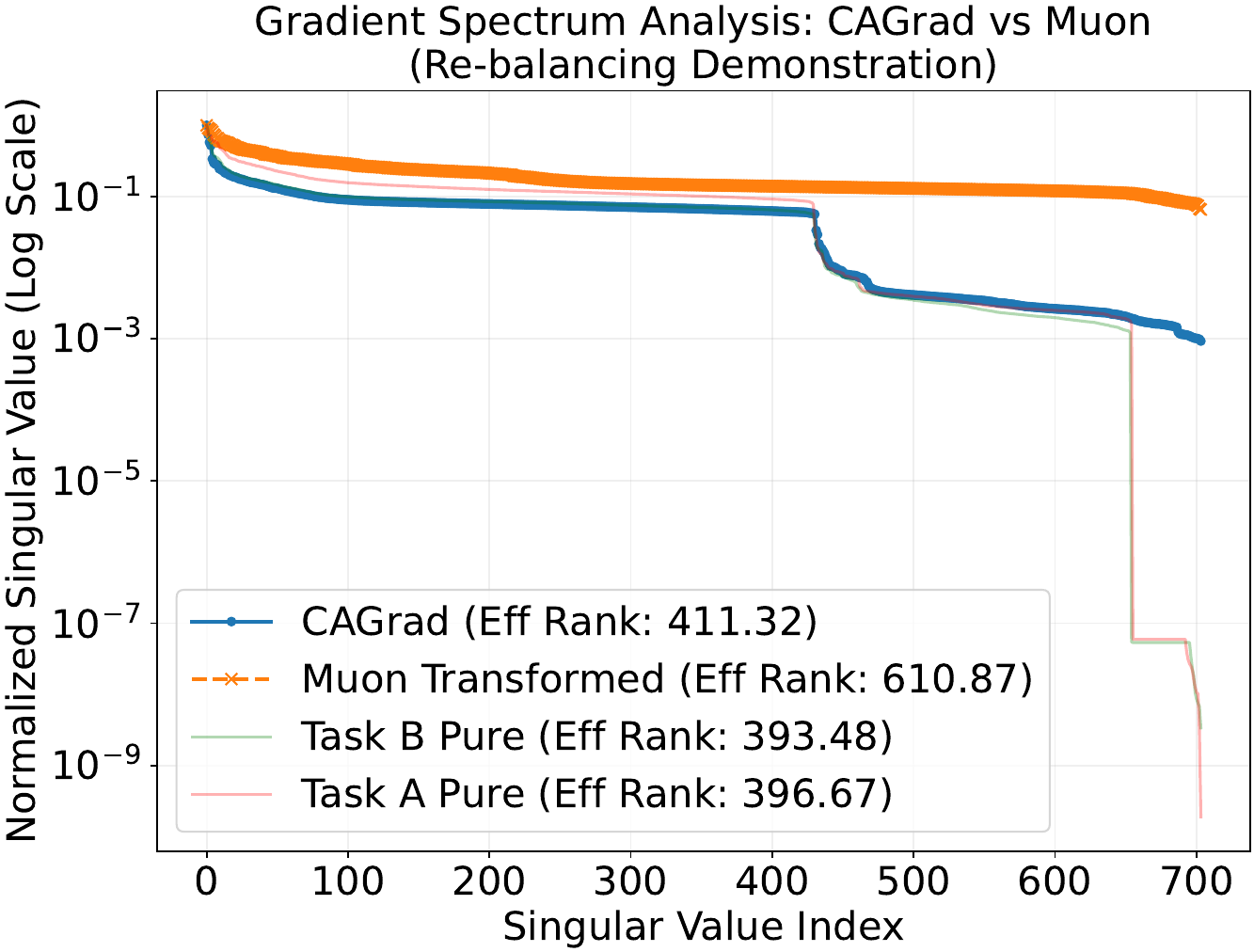}}
    \subfloat[Nash-MTL.]{\includegraphics[width = 0.245\textwidth]{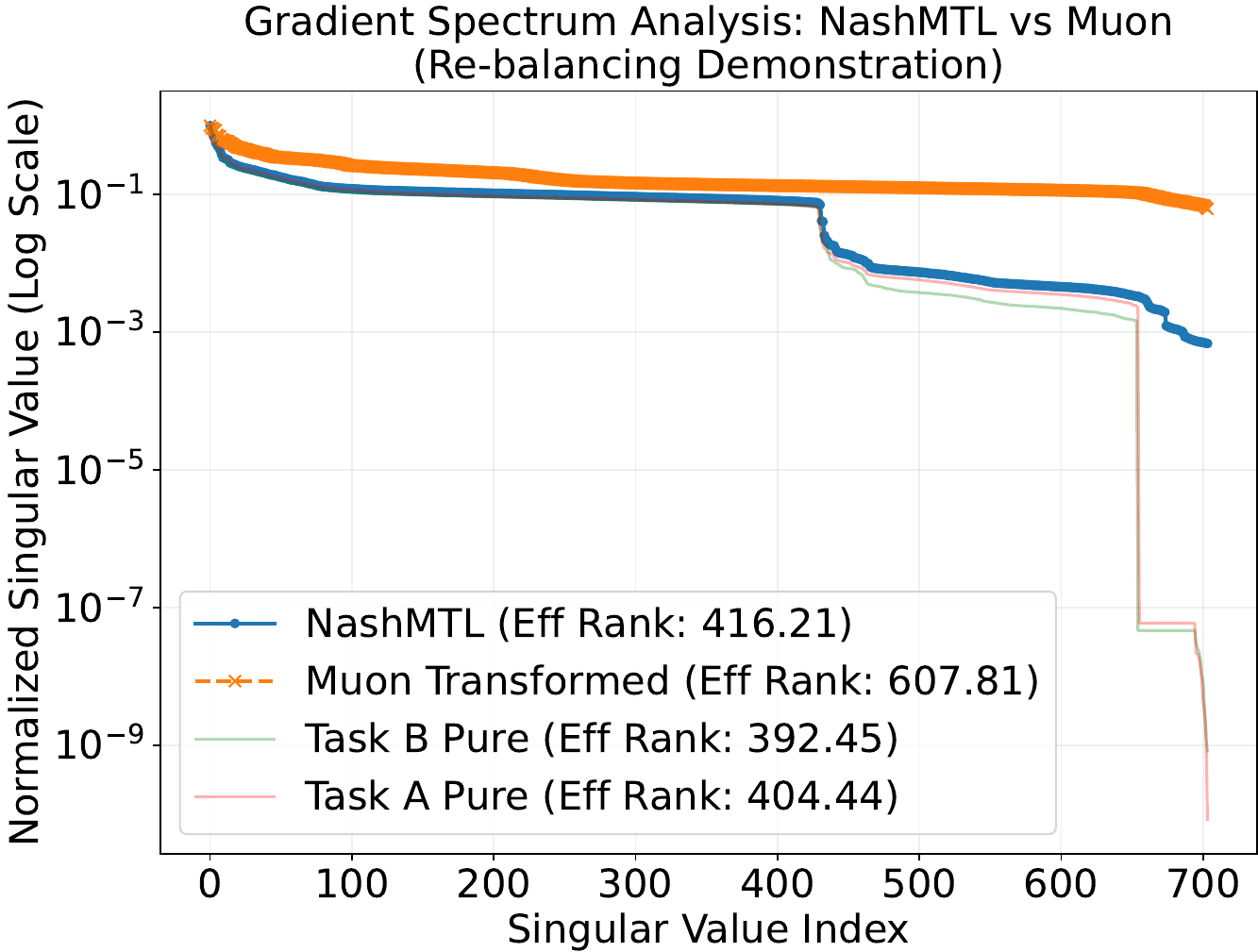}}
    \subfloat[FairGrad.]{\includegraphics[width = 0.245\textwidth]{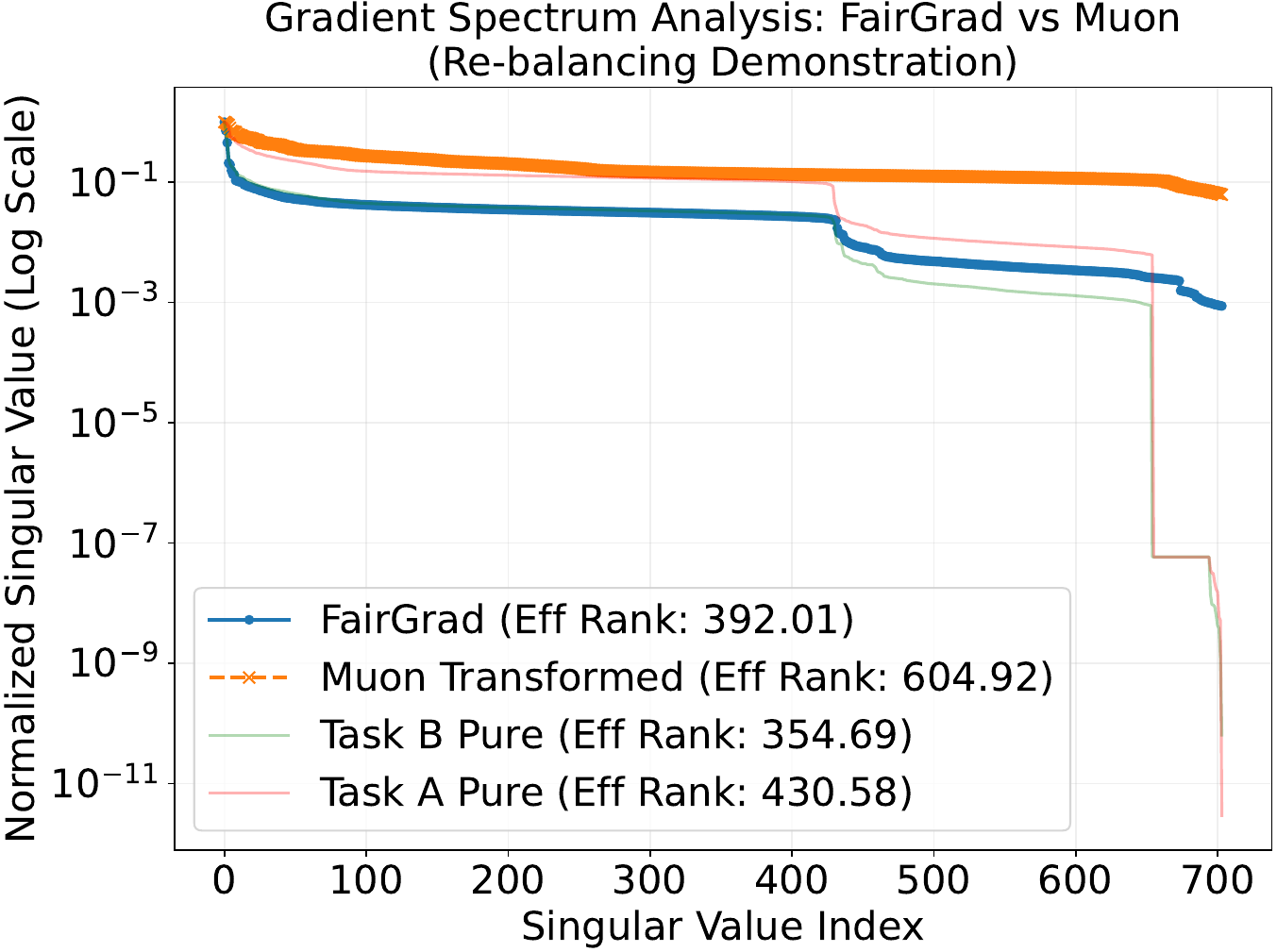}}    
    \caption{Effective rank comparison.} 
    \label{fig:effecive_rank}
\end{figure*}

\subsection{Does MTL Work Like We Expect?} \label{sec:mtl_not_work}
Previous work has rarely examined the impact of EMA–based momentum mechanisms in MTL. Instead, most existing designs implicitly assume a momentum-free optimization setting. In this subsection, we aim to answer a fundamental question: does MTL behave as expected when momentum is introduced?

\noindent \textbf{De-conflicting Is Undermined under Momentum Updates.}
We conduct experiments on CityScapes using FairGrad to investigate how momentum affects the de-conflicting capability of MTL. Specifically, we compute the cosine similarity between the combined gradient $g_{com}$ and each task-specific gradient $\{g_i\}_{i=1}^K$. The results are shown in Figure~\ref{fig:sim_proj}(a)–(c).

For LS, $g_{com}$ is simply the mean of task gradients. For FairGrad, $g_{com}$ is the aggregated gradient produced by the FairGrad algorithm. For Momentum, $g_{com}$ corresponds to the momentum update computed according to Eqn.~\ref{eqn:momentum}, where the instantaneous gradient $G_t$ is obtained from FairGrad.

As illustrated in Figure~\ref{fig:sim_proj}, task gradients under LS are highly imbalanced and can even be conflicting. FairGrad effectively mitigates this issue by producing a combined gradient that is non-conflicting with respect to both tasks. However, when momentum is applied, this favorable property is no longer preserved. The momentum update pulls the final optimization direction toward one that is less aligned with either task gradient, thereby undermining the de-conflicting effect achieved by FairGrad. This observation suggests that naïvely incorporating momentum can negatively interfere with the intended behavior of gradient-based MTL methods.

The empirical evidence presented in Table~\ref{tab:pre_results} on CityScapes, evaluated with four representative MTL approaches—MGDA, CAGrad, Nash-MTL, and FairGrad—further supports our claim. As observed, a lower value of $\beta_1$, which places greater emphasis on the MTL-derived gradient $G_t$ in the parameter update, consistently leads to improved performance across all evaluated methods.

\begin{wrapfigure}[12]{r}{0.48\textwidth}
    \centering
    \subfloat[$\beta_1 = 0.9$.]{\includegraphics[width = 0.24\textwidth]{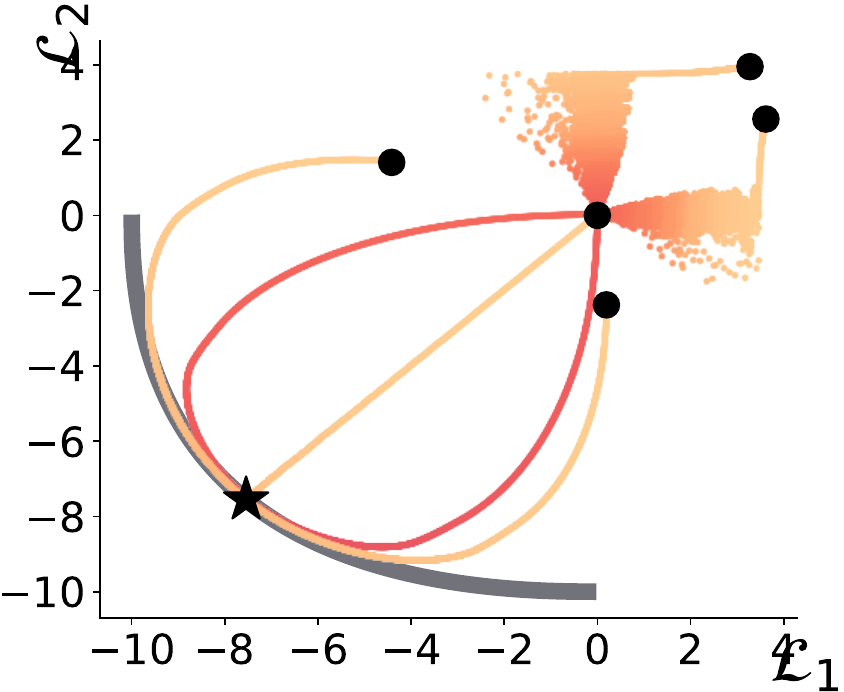}}
    \subfloat[$\beta_1 = 0$.]{\includegraphics[width = 0.24\textwidth]{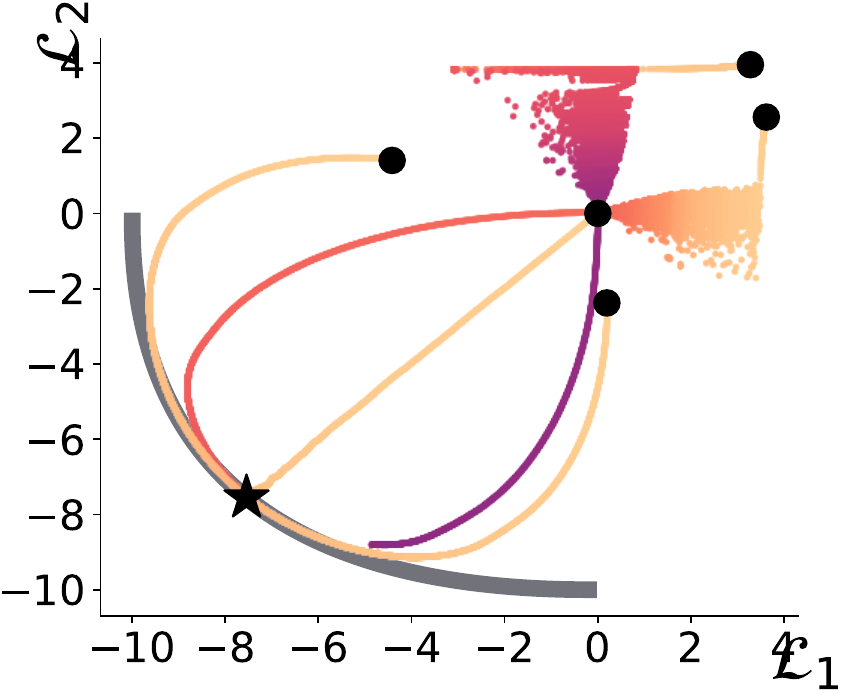}}
    \caption{Effect of momentum on toy example.} 
    \label{fig:toy}
\end{wrapfigure}

\noindent \textbf{Why Current MTL Work?} On the other hand, we emphasize the critical importance of momentum from a multi-task perspective. The EMA mechanism is well-established for its denoising properties, which are vital for stable optimization. Following the toy example in~\cite{liu2023famo}, we compare the optimization trajectories in Figure~\ref{fig:toy} using $\beta_1 = 0.9$ and $\beta_1 = 0$, respectively. As depicted, the trajectory without momentum ($\beta_1 = 0$) is significantly more oscillatory and fails to reach the Pareto front from certain initializations. This demonstrates that while high momentum can undermine instant de-conflicting, the complete absence of momentum sacrifices the stability required to navigate complex multi-task loss landscapes.

Furthermore, as illustrated in Figures~\ref{fig:sim_proj}(a)-(c), while momentum may undermine the immediate de-conflicting capabilities of MTL, it maintains a relatively balanced similarity across all tasks. This suggests that the momentum mechanism effectively mitigates the dominance of any single task. We define this behavior as amortized de-conflicting, where the benefits of MTL are distributed over the optimization timeline rather than applied instantaneously.

Additionally, we compute the projection norms of the final update onto each task gradient and report their projection norm ratios to enable a more rigorous comparison. The results are shown in Figure~\ref{fig:sim_proj}(d). Note that the curves in Figure~\ref{fig:sim_proj}(d) are sampled and smoothed to highlight the overall trend. As observed, MTL methods achieve a clear re-balancing effect compared to the LS baseline. Notably, momentum-based updates exhibit a similar re-balancing behavior, although with more pronounced fluctuations.

\subsection{Muon for MTL}

\noindent \textbf{Muon is an Implicit Multi-Task Learner.} We investigate the theoretical properties of the Muon optimizer in the context of MTL. Muon performs an orthogonalization of the aggregated gradient $G_t$ via the Newton-Schulz iteration, which we then employ for the parameter update:
\begin{align} \label{eqn:muon}
    \theta_{t+1} = \theta_t - \eta O_t, \ \ O_t = \operatorname{Newton-Schulz}(G_t)
\end{align}
The Newton-Schulz iteration approximately computes the polar decomposition $O_t$ of $G_t$, which mathematically corresponds to $O_t = U_tV_t^{\top}$ in the SVD. Consequently, the original gradient $G_t$ can be decomposed as follows:
\begin{align} \label{eqn:svd}
    G_t = U\Sigma V^{\top} = \sum_{k=1}^K \sigma_k u_k v_k^{\top}
\end{align}
where $r = \operatorname{rank}(G_t)$ and the singular values are ordered such that $\sigma_1 \ge \sigma_2 \ge \dots \ge \sigma_r > 0$. We define the orthogonal basis matrices as $E_k = u_k v_k^{\top}$. The projection of each individual task gradient $g_i$ onto this basis is given by:
\begin{align}
    g_i = \sum_{k=1}^r\alpha_{i,k}E_k + G_{i,\bot}, \ \ \alpha_{i,k} = \left \langle g_i, E_k \right \rangle_F 
\end{align}
while the orthogonalized update $O_t$ can be represented as $O_t = \sum_{k=1}^r E_k$.

To analyze the balancing effect, consider the projection of both the standard gradient $G_t$ and the Muon update $O_t$ onto the task gradient $g_i$:
\begin{align}\label{eqn:sgd_proj}
    P_i(G_t) =  \left \langle G_t, g_i \right \rangle_F = \sum_{k=1}^r \sigma_k\alpha_{i,k} \\
    P_i(O_t) = \left \langle O_t, g_i \right \rangle_F = \sum_{k=1}^r \alpha_{i,k}
\end{align}
We define a strong task $i$ as one that contributes a large projection on the primary orthogonal bases and is positively correlated with the singular values $\{\sigma_k\}$, such that $\alpha_{i,1} \ge \alpha_{i,2} \ge \dots \ge \alpha_{i,r} \ge 0$. Conversely, a weak task $j$ contributes less to the primary bases, with the relationship $0 \le \alpha_{j,1} \le \alpha_{j,2} \le \dots \le \alpha_{j,r}$.

Applying Chebyshev's Sum Inequality, we derive the following bounds:
\begin{itemize}
    \item For strong task: \\
    \begin{align}
        P_i(G_t) = \sum_{k=1}^r \sigma_k\alpha_{i,k} \ge \frac{1}{r} \left (  \sum_{k=1}^r \sigma_k\right )\left (  \sum_{k=1}^r \alpha_{i,k}\right ) = \bar{\sigma} P_i(O_t) \nonumber
    \end{align}
    \item For weak task: \\
    \begin{align}
       P_j(G_t) = \sum_{k=1}^r \sigma_k\alpha_{j,k} \le \frac{1}{r} \left (  \sum_{k=1}^r \sigma_k\right )\left (  \sum_{k=1}^r \alpha_{j,k}\right ) = \bar{\sigma} P_j(O_t) \nonumber
    \end{align}
\end{itemize}

Combining these inequalities, we find:
\begin{align}
    \frac{P_i(G_t)}{P_j(G_t)} \ge \frac{\bar{\sigma} P_i(O_t)}{\bar{\sigma} P_j(O_t)} = \frac{P_i(O_t)}{P_j(O_t)}
\end{align}
This result demonstrates that the ratio of task contributions is more balanced in $O_t$ than in $G_t$. Specifically, the orthogonalization process in Muon inherently compresses the dominance of "strong" tasks and amplifies the signal of "weak" tasks. This theoretical finding validates our claim that Muon functions as an implicit multi-task learner by mitigating gradient imbalance.

\noindent \textbf{What Kind of MTL is Suitable for Muon?}
Mathematically, Muon acts as a basis enhancer by orthogonalizing the update matrix, and its effectiveness therefore depends on the directions of the input gradients provided by the MTL algorithm. However, a higher rank is not unconditionally beneficial. Since Muon forces all non-zero singular values to unity ($\sigma_i \to 1$), it may amplify noise by promoting weak or stochastic directions to the same scale as dominant task signals. Consequently, an ideal MTL partner for Muon should preserve a sufficiently high effective rank~\cite{roy2007effective} while allowing Muon to handle the primary balancing through orthogonalization.

As shown in Table~\ref{tab:pre_results}, Muon performs well with CAGrad and Nash-MTL, but less favorably with MGDA and FairGrad. This discrepancy can be attributed to Muon’s spectral orthogonalization, which preserves dominant directions while flattening the singular value spectrum, thereby emphasizing directional geometry over magnitude-based weighting. CAGrad and Nash-MTL naturally align with this mechanism, as they seek central compromise directions and encode task trade-offs mainly through directional structure, resulting in smoother and less degenerate spectra. In contrast, MGDA and FairGrad rely more heavily on spectral magnitude to represent boundary-oriented trade-offs on the Pareto front, making them more sensitive to the loss of magnitude information under spectral flattening.

Further evidence is provided in Figure~\ref{fig:effecive_rank}. FairGrad exhibits the lowest effective rank, while CAGrad and Nash-MTL maintain higher-rank structures. Although MGDA attains a relatively high effective rank, its spectral energy decays more rapidly, further increasing the risk of noise amplification during orthogonalization.

\section{Principal Design}

\subsection{Adaptive Momentum: A Simple Solution}
To tackle \textbf{\textit{Challenge 1}}, we introduce a simple adaptive momentum strategy that dynamically balances instant and amortized de-conflicting based on local curvature.

Intuitively, when the optimization landscape exhibits low curvature, a larger momentum coefficient should be employed to capitalize on temporal filtering and stability while maintaining amortized de-conflicting. Conversely, in regions of high curvature, a lower momentum is preferred to prioritize the immediate resolution of task conflicts.

Following this intuition, we utilize the cosine similarity between gradients of adjacent steps as a proxy for local curvature:
\begin{align} \label{eqn:cos}
    \rho_t = \operatorname{cos}(G_t, G_{t-1})
\end{align}
where $G_t$ is derived MTL algorithm, therefore a high $\rho_t$ indicates a low curvature (similar gradient combination) and a low $\rho_t$ suggests a high curvature (distinct gradient combination). 
We then adaptively adjust the momentum coefficient $\beta_t$ as follows:
\begin{align}
    \beta_t = \beta_{min} + (\beta_{max} - \beta_{min}) \cdot \operatorname{Clip}(\frac{\rho_t + 1}{2}, 0, 1)
\end{align} 
where $\operatorname{Clip}(\cdot, 0, 1)$ ensures the similarity metric remains within a valid range. The hyperparameters $\beta_{max}$ and $\beta_{min}$ define the operational bounds of the momentum, which are usually set as $0.9$ and $0.1$ in our experiments.

\noindent \textbf{Convergence Analysis.} The moment updates are $m_t = \beta_t m_{t-1} + (1 - \beta_t) G_t$ and $v_t = \beta_2 v_{t-1} + (1 - \beta_2) G_t^2$. Incorporating the standard bias correction factors $\rho_{1,t} = 1 - \prod_{j=1}^t \beta_j$ and $\rho_{2,t} = 1 - \beta_2^t$, the parameter update is $\theta_{t+1} = \theta_t - \eta \hat{H}_t^{-1} \hat{m}_t$, where $\hat{H}_t = \text{diag}(\sqrt{\hat{v}_t} + \varepsilon)$.

To facilitate tracking error analysis, we construct an equivalent update step $\theta_{t+1} = \theta_t - \tilde{\eta}_t \hat{H}_t^{-1} m_t$, where $\tilde{\eta}_t = \eta / \rho_{1,t}$ serves as an effective learning rate. Since $\beta_t \le \beta_{max} < 1$, $\tilde{\eta}_t$ is strictly bounded within $[\eta, \tilde{\eta}_{max}]$.

We define the momentum tracking error as $\delta_t = m_t - G_t$, which captures the discrepancy between the historical momentum and the instantaneous MTL gradient. Our first theorem bounds this discrepancy.

\begin{theorem}[Tracking Error Upper Bound]
Under the dynamic update rule, the asymptotic average of the tracking error $\|\delta_t\|^2$ is bounded and explicitly controlled by the dynamic momentum $\beta_t$ and the gradient variation. Specifically, there exists a constant $C > 0$ such that:$$\limsup_{T \to \infty} \frac{1}{T} \sum_{t=1}^T \|\delta_t\|^2 \le C \limsup_{T \to \infty} \frac{1}{T} \sum_{t=1}^T \beta_t^2 \|G_t - G_{t-1}\|^2$$
\end{theorem}

\begin{remark}
Theorem 1 highlights the core mechanism of APT. When task conflicts are severe, the gradient geometry oscillates rapidly, maximizing the variation term $\|G_t - G_{t-1}\|^2$. APT defines $\beta_t = f(\cos(G_t, G_{t-1}))$, which adaptively reduces $\beta_t$ in these exact regions. Consequently, APT dynamically tightens the active upper bound of the tracking error compared to a static, high-momentum baseline.    
\end{remark}

\begin{theorem} [Convergence to Approximate Stationary Neighborhood]
Under the standard assumptions, with a sufficiently small base learning rate $\eta$, the optimization sequence converges to an approximate stationary point of the surrogate objective. The asymptotic convergence neighborhood is tightly bounded by the average tracking error and the inherent conflict residual $\epsilon$:$$\liminf_{T \to \infty} \|\nabla F(\theta_t)\|^2 \le \mathcal{O} \left( \limsup_{T \to \infty} \frac{1}{T} \sum_{t=1}^T \|\delta_t\|^2 \right) + \mathcal{O}(\epsilon)$$
\end{theorem}
\begin{remark}Theorem 2 rigorously bridges the tracking error with the ultimate multi-task optimization goal. A static momentum optimizer suffers from large $\sum \|\delta_t\|^2$ at conflict boundaries, expanding the non-stationary neighborhood. By adaptively suppressing tracking error injection (Theorem 1), APT guarantees a tighter convergence radius around the true Pareto stationary front. 
\end{remark}
 
\subsection{Light Direction Preservation for Muon}
To address \textbf{\textit{Challenge 2}}, we aim to preserve the rich geometric directions of task gradients and maintain a high effective rank, which we have shown to be essential for Muon's orthogonalization process.

To retain the benefits of centered compromise—similar to the objective in CAGrad—we utilize the mean gradient $\bar{g} = \frac{1}{K}\sum_{i=1}^K g_i$ as an anchor direction. We then compute the task-wise cosine similarity between each individual task gradient $g_i$ and the anchor $\bar{g}$:
\begin{align} \label{eqn:task_cos}
    c_i = \operatorname{cos}(g_i, \bar{g})
\end{align}
To ensure the aggregated gradient captures the diverse directional information from all tasks rather than being dominated by a subset, we apply a non-linear gating mechanism:
\begin{align} \label{eqn:gate}
    \omega_i = \operatorname{Sigmoid}(\xi \cdot c_i)
\end{align}
where $\operatorname{Sigmoid}(\cdot)$ denotes the Sigmoid function and $\xi$ is a scaling hyperparameter. The final aggregated gradient is formulated as:
\begin{align}
    G_t = \sum_{i=1}^K \omega_i g_i,\ \ \ \omega_i = K \cdot \frac{\omega_i}{\sum_{i=1}^K\omega_i}
\end{align}
This mechanism serves as a computationally efficient alternative to complex solvers like CAGrad or Nash-MTL. By bypassing the need for numerical iterations or quadratic programming (QP) solvers, it derives a "geometry-rich" input specifically optimized for Muon’s spectral orthogonalization. It is important to note that this approach is a lightweight framework designed to complement Muon, rather than a standalone MTL algorithm. 

\section{Performance Evaluation}
\noindent \textbf{Benchmarks:} We evaluate our method on four widely adopted MTL benchmarks: Cityscapes~\cite{cordts2016cityscapes}, NYUv2~\cite{silberman2012indoor}, CelebA~\cite{liu2015deep}, and QM9~\cite{ramakrishnan2014quantum}. Following the experimental protocols established by FAMO~\cite{liu2023famo}, we categorize these datasets as follows:
\\
\noindent \underline{Scene Understanding}: Cityscapes contains 5,000 annotated street-view images used for two tasks: semantic segmentation and depth estimation. NYUv2 is a three-task benchmark comprising 1,449 annotated indoor images for semantic segmentation, depth estimation, and surface normal prediction.\\
\noindent \underline{Large-scale Classification}: CelebA consists of 200,000 facial images. We treat the classification of 40 binary facial attributes as 40 distinct tasks. \\
\noindent \underline{Graph Regression}: QM9 is a standard benchmark for Graph Neural Network (GNN) research. It includes over 130,000 stable organic molecules represented as graphs with annotated node and edge features. We evaluate this as an 11-task MTL regression problem.

\begin{table*}
\setlength\tabcolsep{3pt}
\centering
\caption{\textbf{Scene understanding} (\textit{NYUv2}, 3 tasks). We report MTAN model performance averaged over 3 random seeds.}
\label{table:nyu}
\footnotesize
\begin{tabular}{llllllllll|ll}
\hline \toprule
\multicolumn{1}{l}{\multirow{3}{*}{Method}} &
  \multicolumn{2}{c}{Segmentation $\uparrow$} &
  \multicolumn{2}{c}{Depth $\downarrow$} &
  \multicolumn{5}{c|}{Surface Normal} &   
  \multicolumn{1}{c}{\multirow{3}{*}{\textbf{MR $\downarrow$}}} &
  \multicolumn{1}{c}{\multirow{3}{*}{\textbf{$\Delta$m\% $\downarrow$}}} \\ \cmidrule(r){2-3} \cmidrule(r){4-5} \cmidrule(r){6-10}
\multicolumn{1}{c}{} &
  \multicolumn{1}{c}{\multirow{2}{*}{mIoU}} &
  \multicolumn{1}{c}{\multirow{2}{*}{Pix. Acc.}} &
  \multicolumn{1}{c}{\multirow{2}{*}{Abs. Err.}} &
  \multicolumn{1}{c}{\multirow{2}{*}{Rel. Err.}} & 
  \multicolumn{2}{c}{Angle Distance $\downarrow$}  &
  \multicolumn{3}{c|}{Within $t^\circ$ $\uparrow$} &
  \multicolumn{1}{c}{} \\ 
\cmidrule(r){6-7} \cmidrule(r){8-10}
\multicolumn{1}{c}{} &
  \multicolumn{1}{c}{} &
  \multicolumn{1}{c}{} &
  \multicolumn{1}{c}{} &
  \multicolumn{1}{c}{} &
  \multicolumn{1}{c}{Mean} &
  \multicolumn{1}{c}{Median} &
  \multicolumn{1}{c}{11.25} &
  \multicolumn{1}{c}{22.5} &
  30 &
  \multicolumn{1}{c}{} \\ \cmidrule(r){1-12} 
Independent &
  \multicolumn{1}{c}{38.30}  &
                     63.76   &                  
  \multicolumn{1}{c}{0.68} &
                     0.28  &                 
  \multicolumn{1}{c}{25.01}  &
  \multicolumn{1}{c}{19.21}  &
  \multicolumn{1}{c}{30.14}  &
  \multicolumn{1}{c}{57.20}  &
                     69.15   &
                     \ \ \ \ -        &
                     \ \ \ \ -
   \\ \cmidrule(r){1-12} 
LS &
  \multicolumn{1}{c}{39.29}  &
                     65.33   &                  
  \multicolumn{1}{c}{0.55} &
                     0.23  &                 
  \multicolumn{1}{c}{28.15}  &
  \multicolumn{1}{c}{23.96}  &
  \multicolumn{1}{c}{22.09}  &
  \multicolumn{1}{c}{47.50}  &
                     61.08   &
                     12.56       &
                     \ 5.46    
   \\
RLW &
  \multicolumn{1}{c}{37.17}  &
                     63.77   &                  
  \multicolumn{1}{c}{0.58} &
                     0.24  &                 
  \multicolumn{1}{c}{28.27}  &
  \multicolumn{1}{c}{24.18}  &
  \multicolumn{1}{c}{22.26}  &
  \multicolumn{1}{c}{47.05}  &
                     60.62   &
                     15.78        &
                     \ 7.67
   \\
DWA &
  \multicolumn{1}{c}{39.11}  &
                     65.31   &                  
  \multicolumn{1}{c}{0.55} &
                     0.23  &                 
  \multicolumn{1}{c}{27.61}  &
  \multicolumn{1}{c}{23.18}  &
  \multicolumn{1}{c}{24.17}  &
  \multicolumn{1}{c}{50.18}  &
                     62.39   &
                     11.78        &
                     \ 3.49
   \\
Uncertainty &
  \multicolumn{1}{c}{36.87}  &
                     63.17   &                  
  \multicolumn{1}{c}{0.54} &
                     0.23  &                 
  \multicolumn{1}{c}{27.04}  &
  \multicolumn{1}{c}{22.61}  &
  \multicolumn{1}{c}{23.54}  &
  \multicolumn{1}{c}{49.05}  &
                     63.65   &
                     12.00        &
                     \ 4.01
   \\
MGDA &
  \multicolumn{1}{c}{30.47}  &
                     59.90   &
  \multicolumn{1}{c}{0.61} &
                     0.26  &
  \multicolumn{1}{c}{24.88}  &
  \multicolumn{1}{c}{19.45}  &
  \multicolumn{1}{c}{29.18}  &
  \multicolumn{1}{c}{56.88}  &
                     69.36   & 
                     9.56        &
                     \ 1.47
   \\ 
GradDrop &
  \multicolumn{1}{c}{39.39}  &
                     65.12   &
  \multicolumn{1}{c}{0.55} &
                     0.23  &
  \multicolumn{1}{c}{27.48}  &
  \multicolumn{1}{c}{22.96}  &
  \multicolumn{1}{c}{23.38}  &
  \multicolumn{1}{c}{49.44}  &
                     62.87   &
                     12.11       &
                     \ 3.61
   \\ 
PCGrad &
  \multicolumn{1}{c}{38.06}  &
                     64.64   &
  \multicolumn{1}{c}{0.56} &
                     0.23  &
  \multicolumn{1}{c}{27.41}  &
  \multicolumn{1}{c}{22.80}  &
  \multicolumn{1}{c}{23.86}  &
  \multicolumn{1}{c}{49.83}  &
                     63.14   &
                     12.67       &
                     \ 3.83
   \\ 
CAGrad &
  \multicolumn{1}{c}{39.79}  &
                     65.49   &
  \multicolumn{1}{c}{0.55} &
                     0.23  &
  \multicolumn{1}{c}{26.31}  &
  \multicolumn{1}{c}{21.58}  &
  \multicolumn{1}{c}{25.61}  &
  \multicolumn{1}{c}{52.36}  &
                     65.58   &
                     9.44       &
                     \ 0.29
   \\ 
IMTL &
  \multicolumn{1}{c}{39.35}  &
                     65.60   &
  \multicolumn{1}{c}{0.54} &
                     0.23  &
  \multicolumn{1}{c}{26.02}  &
  \multicolumn{1}{c}{21.19}  &
  \multicolumn{1}{c}{26.20}  &
  \multicolumn{1}{c}{53.13}  &
                     66.24   &
                     8.33        &
                     -0.59
   \\
Nash-MTL &
  \multicolumn{1}{c}{40.13}  &
                     65.93   &
  \multicolumn{1}{c}{\underline{0.53}} &
                     0.22  &
  \multicolumn{1}{c}{25.26}  &
  \multicolumn{1}{c}{20.08}  &
  \multicolumn{1}{c}{28.40}  &
  \multicolumn{1}{c}{55.47}  &
                     68.15   &
                     5.11        &
                     -4.04
   \\ 
FAMO &
\multicolumn{1}{c}{38.88}  &
                     64.90   &
  \multicolumn{1}{c}{0.55} &
                     0.22  &
  \multicolumn{1}{c}{25.06}  &
  \multicolumn{1}{c}{19.57}  &
  \multicolumn{1}{c}{29.21}  &
  \multicolumn{1}{c}{56.61}  &
                     68.98   &
                     8.56        &
                     -4.10
   \\       
FairGrad &
  \multicolumn{1}{c}{39.74}  &
                     66.01   &
  \multicolumn{1}{c}{0.54} &
                     0.22  &
  \multicolumn{1}{c}{24.84}  &
  \multicolumn{1}{c}{19.60}  &
  \multicolumn{1}{c}{29.26}  &
  \multicolumn{1}{c}{56.58}  &
                     69.16   &
                     \underline{5.00}        &
                     -4.66
   \\ 
COST &
  \multicolumn{1}{c}{38.06}  &
                     64.71   &
  \multicolumn{1}{c}{0.54} &
                     0.23  &
  \multicolumn{1}{c}{\underline{24.47}}  &
  \multicolumn{1}{c}{\cellcolor{mygrey}18.80}  &
  \multicolumn{1}{c}{\underline{30.84}}  &
  \multicolumn{1}{c}{\cellcolor{mygrey}58.25}  &
                     70.30   &
                     \cellcolor{mygrey}4.67       & 
                     -5.39
   \\ 
   \cmidrule(r){1-12} 
\texttt{APT}-CG &
  \multicolumn{1}{c}{\underline{40.51}}  &
                     \underline{66.32}   &
  \multicolumn{1}{c}{0.53} &
                     0.22  &
  \multicolumn{1}{c}{25.77}  &
  \multicolumn{1}{c}{20.65}  &
  \multicolumn{1}{c}{27.24}  &
  \multicolumn{1}{c}{54.25}  &
                     67.15   &
                     6.78   &
                     -2.67$_{\pm0.84}$
   \\   
\texttt{APT}-FAMO &
  \multicolumn{1}{c}{37.66}  &
                     64.57   &
  \multicolumn{1}{c}{0.54} &
                     0.23  &
  \multicolumn{1}{c}{24.87}  &
  \multicolumn{1}{c}{19.11}  &
  \multicolumn{1}{c}{30.12}  &
  \multicolumn{1}{c}{57.50}  &
                     69.58   &
                     7.22 &
                     -4.35$_{\pm2.23}$
   \\  
\texttt{APT}-Fair &
  \multicolumn{1}{c}{38.97}  &
                     {65.07}   &
  \multicolumn{1}{c}{0.54} &
                     \underline{0.22}  &
  \multicolumn{1}{c}{\cellcolor{mygrey}24.44}  &
  \multicolumn{1}{c}{\underline{18.85}}  &
  \multicolumn{1}{c}{\cellcolor{mygrey}30.84}  &
  \multicolumn{1}{c}{\underline{58.12}}  &
                     \cellcolor{mygrey}70.22   &
                     5.67  &
                     \cellcolor{mygrey}-6.04$_{\pm0.22}$
   \\  
\texttt{APT}-Muon &
  \multicolumn{1}{c}{\cellcolor{mygrey}42.47}  &
                     {\cellcolor{mygrey}67.92}   &
  \multicolumn{1}{c}{\cellcolor{mygrey}0.49} &
                     \cellcolor{mygrey}0.22  &
  \multicolumn{1}{c}{25.22}  &
  \multicolumn{1}{c}{20.19}  &
  \multicolumn{1}{c}{28.54}  &
  \multicolumn{1}{c}{55.12}  &
                     67.95 &
                     5.78  &
                     \underline{-5.54$_{\pm0.96}$}
   \\  
   \bottomrule \hline 
\end{tabular}
\end{table*}
\noindent \textbf{Baselines:} We compare it with the following baselines: Linear Scalarization (LS), Scale-Invariant (SI), Random Loss Weighting (RLW) as described in ~\citep{lin2021reasonable}, Dynamic Weight Average (DWA) from ~\citep{liu2019end}, Uncertainty Weighting (UW) detailed in ~\citep{kendall2018multi}, MGDA from ~\citep{sener2018multi}, GradDrop presented in ~\citep{chen2020just}, PCGrad as in ~\citep{yu2020gradient}, CAGrad from ~\citep{liu2021conflict}, IMTL detailed in ~\citep{liu2021towards}, Nash-MTL from ~\citep{navon2022multi}, FAMO described in ~\citep{liu2024famo}, FairGrad from ~\citep{ban2024fair}, and COST from~\cite{zhou2025continual}.

\noindent \textbf{Evaluation Metric:} In addition to individual task performance, we incorporate a widely used aggregate metric, $\Delta m\%$~\citep{maninis2019attentive}. This metric evaluates the overall performance degradation relative to independently trained models, which serve as reference oracles. The formal definition of $\bm{\Delta m\%}$ is:
\begin{align} \label{eqn:deltam}
\Delta m\% = \frac{1}{K} \sum_{k=1}^K (-1)^{\delta_k} \frac{(M_{m,k} - M_{b,k})}{M_{b,k}} \times 100
\end{align}
\begin{wraptable}[25]{r}{0.54\textwidth}
\vspace*{-\baselineskip}
\setlength\tabcolsep{2pt}
\centering
\caption{\textbf{Scene understanding} (\textit{CityScapes}, 2 tasks). We report MTAN model performance averaged over 3 random seeds. The best scores are provided in \colorbox{gray!20}{gray}, and the second scores are \underline{underlined}.}
\label{table:city}
\footnotesize
\begin{tabular}{lllll|ll}
\hline \toprule
\multicolumn{1}{l}{\multirow{2}{*}{Method}} &
  \multicolumn{2}{c}{Segmentation $\uparrow$} &
  \multicolumn{2}{c|}{Depth $\downarrow$} &   
  \multicolumn{1}{c}{\multirow{2}{*}{\textbf{MR $\downarrow$}}} &
  \multicolumn{1}{c}{\multirow{2}{*}{\textbf{$\Delta$m\% $\downarrow$}}} \\ \cmidrule(r){2-3} \cmidrule(r){4-5}
  
\multicolumn{1}{c}{} &
  \multicolumn{1}{c}{mIoU} &
  \multicolumn{1}{c}{Pix.} &
  \multicolumn{1}{c}{Abs.} &
  \multicolumn{1}{c|}{Rel.} &
  \multicolumn{1}{c}{} \\ \cmidrule(r){1-7} 
Independent &
  \multicolumn{1}{c}{74.01}  &
  \multicolumn{1}{c}{93.16}   &                  
  \multicolumn{1}{c}{0.0125} &
  \multicolumn{1}{c|}{27.77}  &                 
                     -      &                 
                     -
   \\ \cmidrule(r){1-7}  
LS &
  \multicolumn{1}{c}{75.18}  &
  \multicolumn{1}{c}{93.49}   &
  \multicolumn{1}{c}{0.0155} &
                     46.77  &
                     11.25       &
                     22.60
   \\ 
RLW &
  \multicolumn{1}{c}{74.57}  &
  \multicolumn{1}{c}{93.41}   &
  \multicolumn{1}{c}{0.0158} &
                     47.79  &
                     14.50      &
                     24.37
   \\ 
DWA &
  \multicolumn{1}{c}{75.24}  &
  \multicolumn{1}{c}{93.52}   &
  \multicolumn{1}{c}{0.0160} &
                     44.37  &
                     11.25         &
                     21.43
   \\ 
Uncertainty &
  \multicolumn{1}{c}{72.02}  &
  \multicolumn{1}{c}{92.85}   &
  \multicolumn{1}{c}{0.0140} &
                     30.13  &
                     10.75        &
                     5.88
   \\ 
MGDA &
  \multicolumn{1}{c}{68.84}  &
  \multicolumn{1}{c}{91.54}   &
  \multicolumn{1}{c}{0.0309} &
                     33.50  &
                     14.50       &
                     44.14
   \\ 
GradDrop &
  \multicolumn{1}{c}{75.27}  &
  \multicolumn{1}{c}{93.53}   &
  \multicolumn{1}{c}{0.0157} &
                     47.54  &
                     10.75       &
                     23.67
   \\ 
PCGrad &
  \multicolumn{1}{c}{75.13}  &
  \multicolumn{1}{c}{93.48}   &
  \multicolumn{1}{c}{0.0154} &
                     42.07  &
                     11.50        &
                     18.21
\\
CAGrad &
  \multicolumn{1}{c}{75.16}  &
  \multicolumn{1}{c}{93.48}   &
  \multicolumn{1}{c}{0.0141} &
                     37.60  &
                     10.25        &
                     11.58
   \\
IMTL &
  \multicolumn{1}{c}{75.33}  &
  \multicolumn{1}{c}{93.49}   &
  \multicolumn{1}{c}{0.0135} &
                     38.41  &
                     8.50      &
                     11.04
   \\  
Nash-MTL &
  \multicolumn{1}{c}{75.41}  &
  \multicolumn{1}{c}{93.66}   &
  \multicolumn{1}{c}{0.0129} &
                     35.02  &
                     4.75        &
                     6.82
   \\  
FAMO &
  \multicolumn{1}{c}{74.54}  &
  \multicolumn{1}{c}{93.29}   &
  \multicolumn{1}{c}{0.0145} &
                     32.59  &
                     11.25       &
                     8.13
   \\  
FairGrad &
  \multicolumn{1}{c}{75.72}  &
  \multicolumn{1}{c}{\underline{93.68}}   &
  \multicolumn{1}{c}{0.0134} &
                     32.25 &
                     \underline{4.25}       & 
                     5.18 
   \\   
COST &
  \multicolumn{1}{c}{75.73}  &
  \multicolumn{1}{c}{93.53}   &
  \multicolumn{1}{c}{0.0133} &
                     31.53 &
                     \underline{4.25}       &
                     4.30
   \\ 
    \cmidrule(r){1-7} 
\texttt{APT}-CG &
  \multicolumn{1}{c}{\underline{75.79}}  &
  \multicolumn{1}{c}{93.65}   &
  \multicolumn{1}{c}{0.0129} &
                     38.13  &
                     5.25      &
                     9.31$_{\pm1.21}$ \\    
\texttt{APT}-FAMO &
  \multicolumn{1}{c}{75.11}  &
  \multicolumn{1}{c}{93.48}   &
  \multicolumn{1}{c}{0.0138} &
                     35.39  &
                    10.25     &
                     9.03$_{\pm1.23}$ \\   
\texttt{APT}-Fair &
  \multicolumn{1}{c}{74.29}  &
  \multicolumn{1}{c}{93.25}   &
  \multicolumn{1}{c}{\underline{0.0125}} &
                     30.36 &
                    8.75     &
                     \underline{2.22$_{\pm1.46}$} \\   
\texttt{APT}-Muon &
  \multicolumn{1}{c}{\cellcolor{mygrey}78.23}  &
  \multicolumn{1}{c}{\cellcolor{mygrey}94.36}   &
  \multicolumn{1}{c}{\cellcolor{mygrey}0.0121} &
                     \cellcolor{mygrey}27.09  &
                    \cellcolor{mygrey}1.00       &
                     \cellcolor{mygrey}-3.09$_{\pm1.16}$ \\   
 \bottomrule \hline
\end{tabular}
\end{wraptable}
where $M_{m,k}$ and $M_{b,k}$ represent the value of the $k$-th metric for the compared multi-task method and the independent single-task baseline, respectively. The indicator $\delta_k$ is set to 1 if a higher value is preferred for metric $M_k$, and 0 otherwise. Furthermore, we report the Mean Rank (\textbf{MR}), which calculates the average ranking of each method across all tasks; a lower MR signifies more consistent performance across the benchmark.

\noindent \textbf{Implementation Details:} Adam optimizer settings. For the Cityscapes and NYUv2 datasets, we construct our model using the SegNet architecture~\cite{badrinarayanan2017segnet} and incorporate MTAN~\cite{liu2019end} as the shared backbone. We train the models for 200 epochs using the Adam optimizer with an initial learning rate of $1.0 \times 10^{-4}$, which is halved after the 100th epoch. The batch size is set to 8 for Cityscapes and 2 for NYUv2. For CelebA, we utilize a 9-layer Convolutional Neural Network (CNN) as the backbone, followed by task-specific linear heads. The model is trained for 15 epochs with the Adam optimizer, using a learning rate of $3.0 \times 10^{-4}$ and a batch size of 256. For QM9, our model is trained for 300 epochs with a batch size of 120. We initialize the learning rate at $1.0 \times 10^{-3}$ and employ a "Reduce On Plateau" scheduler, which decreases the learning rate when validation performance stagnates, ensuring consistent convergence with prior work.

Muon Optimizer Settings. Regarding the Muon optimizer, we maintain consistency by sharing all architectural and training configurations with our Adam optimizer setup, with the exception of the learning rate. For CityScapes and NYUv2, the learning rate is set as 0.02; For CelebA, the learning rate is set as 5.e-3; For QM9, the learning rate is initialized as 5.e-3. We employ a "Reduce On Plateau" scheduler with a patience of 10 steps, decreasing the learning rate once the validation performance stagnates. 

\subsection{Toy Example} 
\begin{figure*}
    \centering
    \subfloat[LS]{\includegraphics[width = 0.21\textwidth]{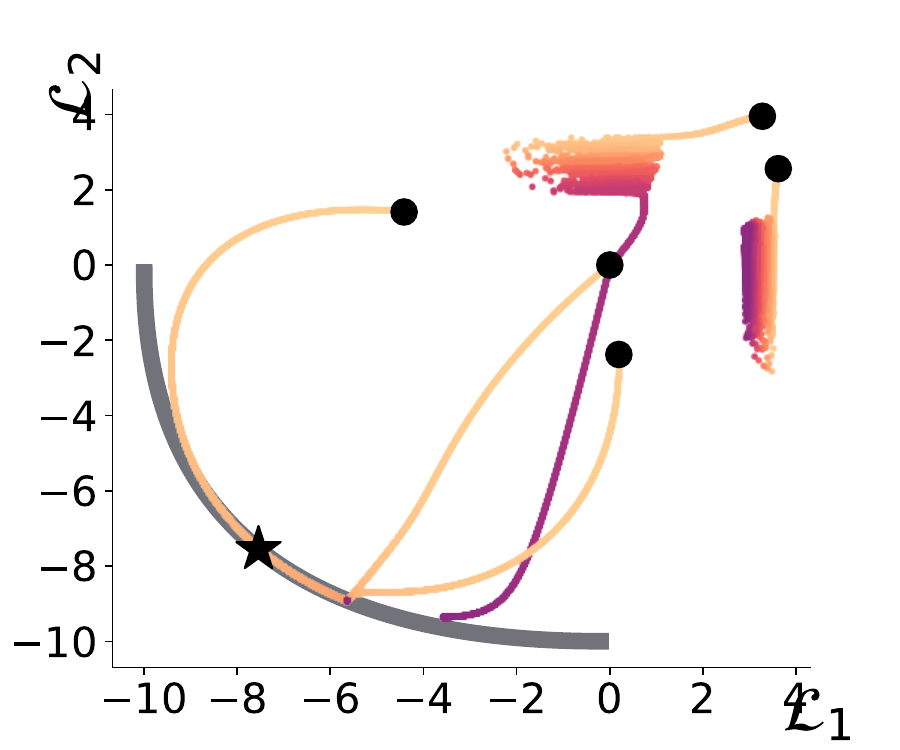}}
    \subfloat[CAGrad]{\includegraphics[width = 0.19\textwidth]{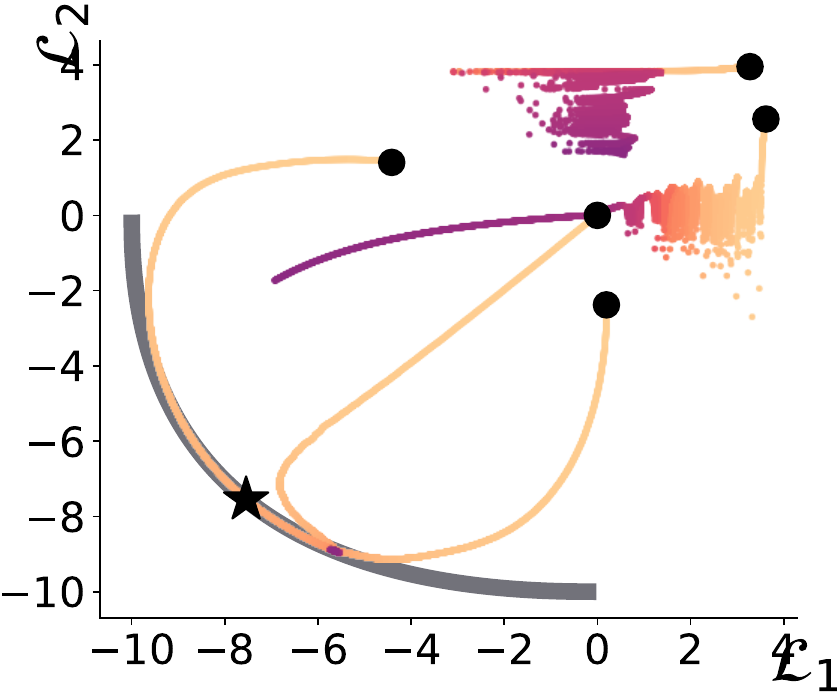}}
    \subfloat[\texttt{APT}-CAGrad]{\includegraphics[width = 0.19\textwidth]{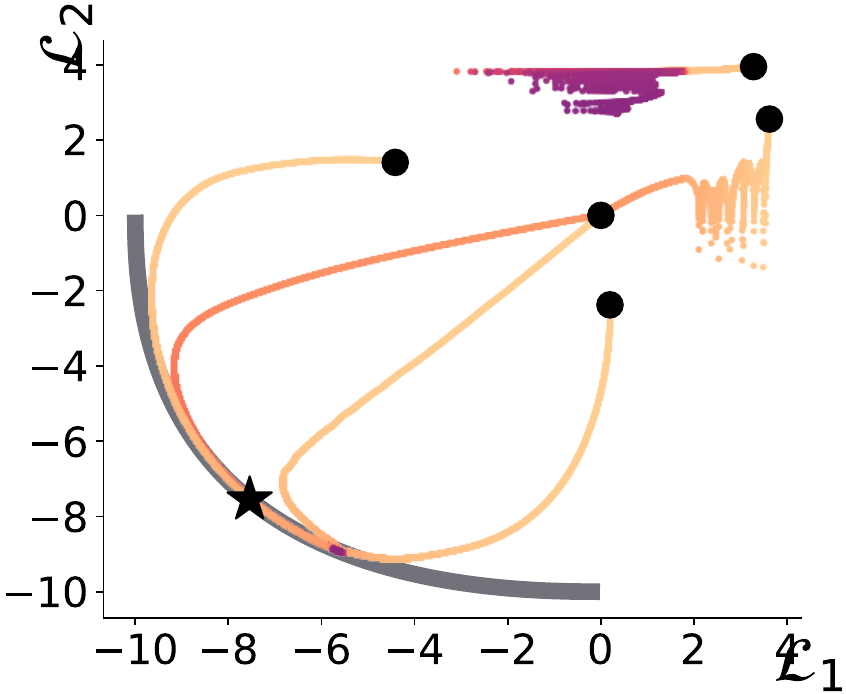}}
    \subfloat[FairGrad]{\includegraphics[width = 0.19\textwidth]{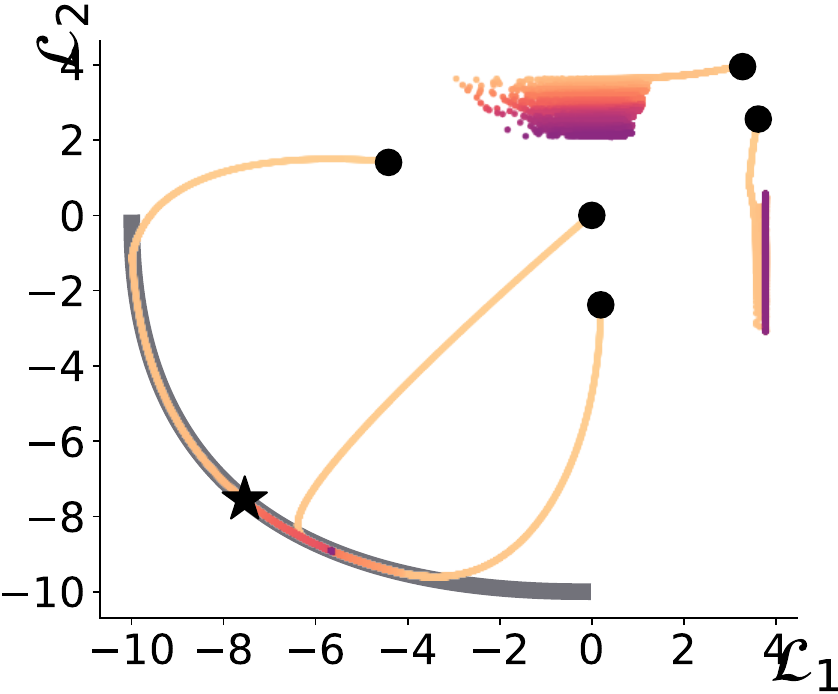}}    
    \subfloat[\texttt{APT}-FairGrad]{\includegraphics[width = 0.19\textwidth]{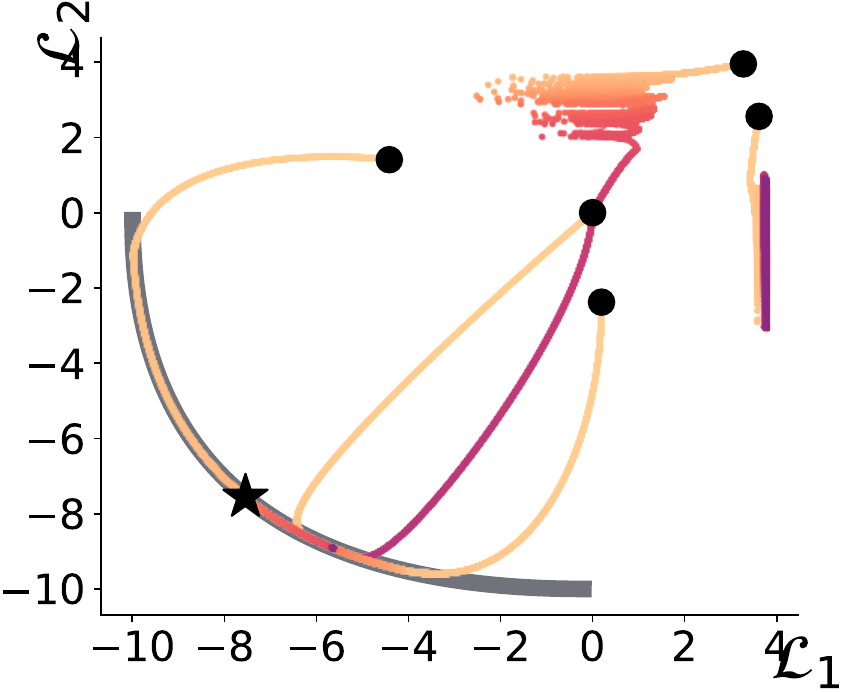}}    
    \caption{Toy example comparison.} 
    \label{fig:toy_eva}
\end{figure*}
To further validate the effectiveness of our approach, we provide comparisons in Figure~\ref{fig:toy_eva}. In this evaluation, we select CAGrad and FairGrad as representative baselines and augment them with the \texttt{APT} framework. As illustrated, the vanilla baselines struggle with certain initializations, failing to reach the Pareto front. In contrast, the integration of \texttt{APT} significantly enhances their convergence, guiding the less noisy optimization trajectory toward the Pareto front.
\begin{wraptable}[19]{r}{0.45\textwidth}
\vspace*{-1.0\baselineskip}
\setlength\tabcolsep{3pt}
\centering
\caption{Results on \textit{CelebA} and \textit{QM9} datasets with MR and $\Delta$m\%.}
\label{table:qm9}
\footnotesize
\begin{tabular}{lcc|cc}
\hline \toprule 
\multirow{2}{*}{\textbf{Method}} & \multicolumn{2}{c}{\textbf{CelebA}} & \multicolumn{2}{c}{\textbf{QM9}} \\
\cmidrule(lr){2-3} \cmidrule(lr){4-5} 
 & \textbf{MR $\downarrow$} & \textbf{$\Delta$m\% $\downarrow$} & \textbf{MR $\downarrow$} & \textbf{$\Delta$m\% $\downarrow$} \\
\midrule
LS & 9.35 & 4.15 & 12.45 & 177.6 \\
SI & 11.63 & 7.20 & 7.36 & 77.8 \\
RLW & 8.13 & 1.46 & 14.27 & 203.8 \\
DWA & 10.40 & 2.40 & 12.27 & 175.3 \\
UW & 8.70 & 3.23 & 9.09 & 108.0 \\
MGDA & 15.68 & 14.85 & 12.09 & 120.5 \\
PCGrad & 10.03 & 3.17 & 10.73 & 125.7 \\
CAGrad & 9.68 & 2.48 & 11.45 & 112.8 \\
IMTL-G & 7.50 & 0.84 & 10.00 & 77.2 \\
Nash-MTL & 7.68 & 2.84 & 6.00 & 62.0 \\
FAMO & 7.50 & 1.21 & 7.91 & 58.5 \\
FairGrad & 8.15 & \underline{0.37} & \underline{5.82} & \underline{57.9} \\
COST & 9.03 & 0.93 & 6.82 & 58.3 \\ 
\midrule
\texttt{APT}-CG & 7.30 & 1.12$_{\pm1.39}$ & 9.73 & 108.9$_{\pm1.57}$ \\
\texttt{APT}-FAMO & \underline{7.20} & 1.04$_{\pm0.64}$ & 6.64 & \cellcolor{mygrey}55.5$_{\pm7.79}$ \\
\texttt{APT}-Fair & \cellcolor{mygrey}6.75 & 0.53$_{\pm1.55}$ & \cellcolor{mygrey}4.36 & 64.9$_{\pm3.55}$ \\
\texttt{APT}-Muon & 8.33 & \cellcolor{mygrey}0.03$_{\pm0.62}$ & 6.00 & 72.9$_{\pm1.63}$ \\
\bottomrule \hline
\end{tabular}
\end{wraptable}

\subsection{Main Results}
We evaluate our framework on the four aforementioned Multi-Task Learning (MTL) benchmarks. Specifically, we augment representative MTL approaches—including CAGrad (denoted as CG), FAMO, and FairGrad (denoted as Fair)—with the \texttt{APT} framework. We also include a Muon-based variant, \texttt{APT}-Muon.

The results for the scene understanding benchmarks are presented in Table~\ref{table:city} and Table~\ref{table:nyu}. As shown, \texttt{APT} generally augments all evaluated baselines, yielding considerable improvements in both Mean Rank (MR) and $\Delta m\%$. Notably, \texttt{APT}-Muon consistently attains superior performance on both benchmarks. This demonstrates that Muon inherently functions as an effective MTL learner, while our proposed framework helps preserve the diversity of gradient directions necessary for stable optimization. Additionally, \texttt{APT}-Fair exhibits near-SOTA performance, further highlighting the critical importance of balancing instant and amortized de-conflicting.
In many-task scenarios such as CelebA (40 tasks) and QM9 (11 tasks), \texttt{APT} continues to demonstrate strong augmentation effects across most baselines. While we observe a slight performance dip when paired with FairGrad on CelebA—which may stem from the increased complexity of balancing 40 distinct tasks—\texttt{APT}-Muon maintains its impressive SOTA standing and remains highly competitive on QM9. Furthermore, \texttt{APT}-FAMO achieves SOTA results on the QM9 benchmark.

These results collectively validate the effectiveness of our proposed framework, demonstrating that \texttt{APT} is both a simple and highly effective solution for enhancing modern MTL optimization. 



\section{Conclusion}
In this work, we revisited the effectiveness of optimization-based MTL and identified a critical bottleneck: the disconnect between derived task gradients and the actual parameter updates. Our analysis revealed that this discrepancy prevents previous MTL frameworks from fully leveraging their intended learning dynamics. Furthermore, we highlighted the intrinsic multi-task capabilities of the Muon optimizer and established the importance of high-quality gradient inputs for its orthogonalization process. To bridge these gaps, we introduced the \texttt{APT} framework, which integrates a simple adaptive momentum mechanism with a light direction preservation strategy. Extensive empirical evaluations across four mainstream datasets confirm that \texttt{APT} successfully harmonizes advanced optimizers with MTL. 

\bibliographystyle{plainnat}
\bibliography{example_paper}

\end{document}